\documentclass{article}
\usepackage[preprint]{log_2024}					

\usepackage{booktabs}	
\usepackage{multirow}		
\usepackage{amsfonts}
\usepackage{graphicx}						

\usepackage[numbers,compress,sort]{natbib}

\title[Transfer Learning for Temporal Link Prediction]{Transfer Learning for Temporal Link Prediction}

\author[tlp]{
Ayan Chatterjee \\
BioClarity AI \\
Northeastern University \\
\email{bioclarityai@gmail.com} \And
Barbara Ikica \\
Google \\
\email{barbaraikica@google.com} \And
Babak Ravandi \\
Alexion AstraZeneca Rare Disease \\ Northeastern University \\
\email{bk.ravandi@gmail.com} \And
John Palowitch \\
Google DeepMind \\
\email{palowitch@google.com}
}

\begin{document}

\maketitle

\begin{abstract}
Link prediction on graphs has applications spanning from recommender systems to drug discovery. Temporal link prediction (TLP) refers to predicting future links in a temporally evolving graph and adds additional complexity related to the dynamic nature of graphs. State-of-the-art TLP models incorporate memory modules alongside graph neural networks to learn both the temporal mechanisms of incoming nodes and the evolving graph topology. However, memory modules only store information about nodes seen at train time, and hence such models cannot be directly transferred to entirely new graphs at test time and deployment. In this work, we study a new transfer learning task for temporal link prediction, and develop transfer-effective methods for memory-laden models. Specifically, motivated by work showing the informativeness of structural signals for the TLP task, we augment a structural mapping module to the existing TLP model architectures, which learns a mapping from graph structural (topological) features to memory embeddings. Our work paves the way for a memory-free foundation model for TLP.
\end{abstract}

\section{Introduction}
Link prediction is a fundamental task in the field of machine learning and network science, focusing on predicting the unobserved connections in graphs \cite{feng2023surveylinkpredictionalgorithms,10.1145/3012704}. 
Temporal link prediction (TLP) refers to link prediction in graphs where nodes and edges evolve over time \cite{Divakaran2019,10.1145/3625820}. 
The task spans a wide range of real-world graphs, including social networks \cite{9378444}, biological systems \cite{Hosseinzadeh2022}, and financial transactions \cite{kim2024temporalgraphnetworksgraph}. 
The ability to accurately predict the emergence of links over time is crucial for tasks such as recommendation systems \cite{10.1145/2063576.2063744}, anomaly detection \cite{anonymous2024learningbased}, and epidemic modeling \cite{PhysRevX.9.031017}.

Recent research \cite{Chatterjee2023,chatterjee2023inductive,Chatterjee2024} has used unsupervised pre-training of node attributes to address inductive link prediction in both static and temporal graphs. However, practical constraints often render node attribute data unavailable, particularly due to confidentiality concerns in social networks \cite{Rajaei2015} and biological/medical networks \cite{Su2021}. Furthermore, going beyond inductive tests and transferring a link prediction model to an entirely distinct graph poses challenges such as aligning latent spaces \cite{9679004}. The proposed solution of augmentation of loss functions by Gritsekno et al. does not truly emulate a transfer learning scenario \cite{9679004}. The complexity further escalates in temporal graphs, where transferring both spatial and temporal mechanisms to a new graph becomes necessary.

In this paper, we propose a novel approach to temporal link prediction that addresses these challenges. Our method incorporates topological features of the nodes, which allows for a common space between two disjoint graphs. We evaluate our approach on several benchmark datasets, demonstrating its effectiveness and efficiency compared to state-of-the-art methods.

The rest of this paper is organized as follows. In Section 2, we review related work on temporal link prediction. Section 3 introduces the problem definition and summarizes the limitations of the existing TLP models in transfer learning. Section 4 introduces our proposed method, followed by experimental results in Section 5. Finally, we conclude with the limitations, a discussion, and future work in Sections 6 and 7.

\section{Related Work}

A temporal graph is a graph in which new nodes and new edges arrive over time. The standard temporal
link prediction (TLP) task is to predict future edges between nodes that currently exist in the graph,
constituting a transductive link prediction scenario. 
In static graphs, link prediction necessitates capturing the neighborhood information of individual nodes. 
This is achieved by neighborhood aggregation \cite{graphsage,grail}. Conversely, the temporal domain adds an additional layer of complexity by requiring learning the sequential processes governing link formation \cite{10.1145/3625820}. 
While temporally mature dynamic graphs \cite{10.1145/2872518.2889398} offer sufficient neighborhood topology for transductive link prediction with high accuracy, forecasting edges for newly introduced nodes in an inductive setting presents a formidable challenge. 
Many SOTA temporal link prediction models, such as TGAT \cite{Xu2020Inductive} and DNformer \cite{10.1145/3551892}, can tackle the inductive scenario predicting links between new nodes by
aggregating the induced subgraph of an unseen node containing seen nodes.

However, the case of a test graph with a completely new set of nodes with no existing subgraphs to aggregate has yet to be considered. Such transfer tasks have been considered in static graphs. 
A graph coupling approach has been proposed by Gritsenko et al. \cite{Gritsenko2022} for transfer learning in
node classification by aligning the node embedding spaces. However, the approach requires simultaneous observation of both train and test graphs, which is impractical for evolving graphs, and also
constitutes an unrealistic transfer learning scenario in which the test graph is seen during training. 

Latent spaces are crucial in temporal graph learning as they encode graph nodes, edges, and dynamic interactions into lower-dimensional representations while retaining significant structural and temporal information. These representations capture temporal dependencies and variations in node relationships over time, enabling the efficient prediction of future states, interactions, or events. Temporal graph learning frameworks leverage latent spaces to map complex temporal patterns into features that can be processed by predictive models. For instance, graph neural networks (GNNs) equipped with temporal mechanisms, such as temporal attention or recurrent layers, project dynamic graph elements into latent spaces where time-dependent patterns are disentangled from static structures \cite{Kipf:2017tc,Rossi_8519335}. This abstraction facilitates downstream tasks like temporal link prediction and anomaly detection, allowing for generalization across varying temporal dynamics \cite{Xu2020Inductive}.

Recently, the graph machine learning community has focused on the need to develop static graph foundation models (GFMs) \cite{mao2024graph}. However, transferability in temporal graphs remains unexplored and is a key step towards developing a temporal graph foundation model. Graph foundation models aim to generalize across diverse graph-based tasks, including temporal graph learning, by pretraining on extensive graph datasets and fine-tuning them for specific applications. These models often incorporate temporal encoding mechanisms to process dynamic changes in graph topology and node interactions. GFMs leverage self-supervised pretraining tasks like masked edge prediction or temporal node classification to capture both temporal evolution and structural dependencies within dynamic graphs \cite{10.1145/3589335.3651980}. Recent innovations integrate time-aware embeddings and memory-augmented structures to enhance temporal generalization capabilities. By adopting pretraining paradigms common in natural language processing and computer vision, GFMs provide robust latent representations that can adapt to various temporal graph learning challenges, such as forecasting time-evolving relationships in social networks or predicting drug-target interactions in dynamic biomedical graphs \cite{graphsage,Jin2020SelfsupervisedLO}.

\section{Problem Definition}

\subsection{Preliminaries}

Static graphs are usually modeled as a tuple $G = (V, E, f, w)$, where $V$ is the set of nodes, $E \subseteq V \times V$ is the set of edges, $f : V \rightarrow R^{d_N}$ are the node features and $w : E \rightarrow R^
{d_E}$ are the edge features. Dynamic graphs are more complex to represent, as it is necessary to capture their evolution over time. In this work, we focus on the commonly used Continuous-Time Dynamic Graph (CTDG) representation \cite{Nguyen2018}. 
A CTDG is a tuple $(G(0), T)$, where $G(0) = (V(0), E(0), f(0), w(0))$ represents the initial state of the graph and $T$ is a set of tuples
of the form (timestamp, event) representing events to be applied to the graph at given timestamps.
These events could be node additions/deletions, edge additions/deletions, or feature updates.
Assume a CTDG $(G(0), T)$, a timestamp $t \in N$, and an edge $(u, v)$. In dynamic link prediction, the task is to decide, whether there is some event $e$ pertaining to $(u, v)$, such that $(t, e) \in T$, while only considering the CTDG $(G(0), { (t^{'}, e) \in T | t^{'} < t })$. This link prediction problem is usually
considered in three settings, the transductive setting (predicting links between nodes that were seen
during training), the semi-inductive setting (predicting links between two nodes, one of the nodes seen during training and the other unseen during training), and the inductive setting (predicting links between nodes not seen during training). In this work, we focus on link addition and deletion in an inductive setting, where the test nodes are completely disjoint from the train nodes and pertain to a graph different from the train graph. 

\subsection{Background: Temporal Graph Networks}

This paper studies the limitations of \textbf{Temporal Graph Networks (TGNs)} \cite{rossi2020temporal}, a class of deep learning models designed to learn dynamic node representations for both link prediction and node classification. Unlike static graph neural networks (GNNs) that operate on fixed graph structures, TGN captures temporal dependencies by incorporating memory mechanisms, message passing, time encoding, and a GNN-based aggregation step. Specifically, they consist of the following components: (1) a \textbf{memory module} that maintains temporal embeddings for each node, continuously updating them as new interactions occur; (2) a \textbf{message function} that aggregates information from neighboring nodes and edges, allowing the model to incorporate temporal dependencies in message passing; (3) a \textbf{memory updater}, typically implemented as a gated recurrent unit (GRU) or an MLP, which refines node memories based on new interactions; (4) a \textbf{temporal encoder}, which models the time gaps between interactions to capture temporal patterns effectively; and finally, (5) a downstream \textbf{GNN readout} module, which learns a node embedding function that propagates the memory/message embeddings across the graph, typically using attention-based or graph convolution-based architectures to enhance structural learning. TGN models with these five components as a backbone can be trained for temporal link prediction (presence/absence of future edges) and temporal node property prediction (when a node will adopt/lose a certain state).

\subsection{Limitations of TGN Memory Embeddings}

\begin{figure*}
	\centering
	\includegraphics[width=1\linewidth]{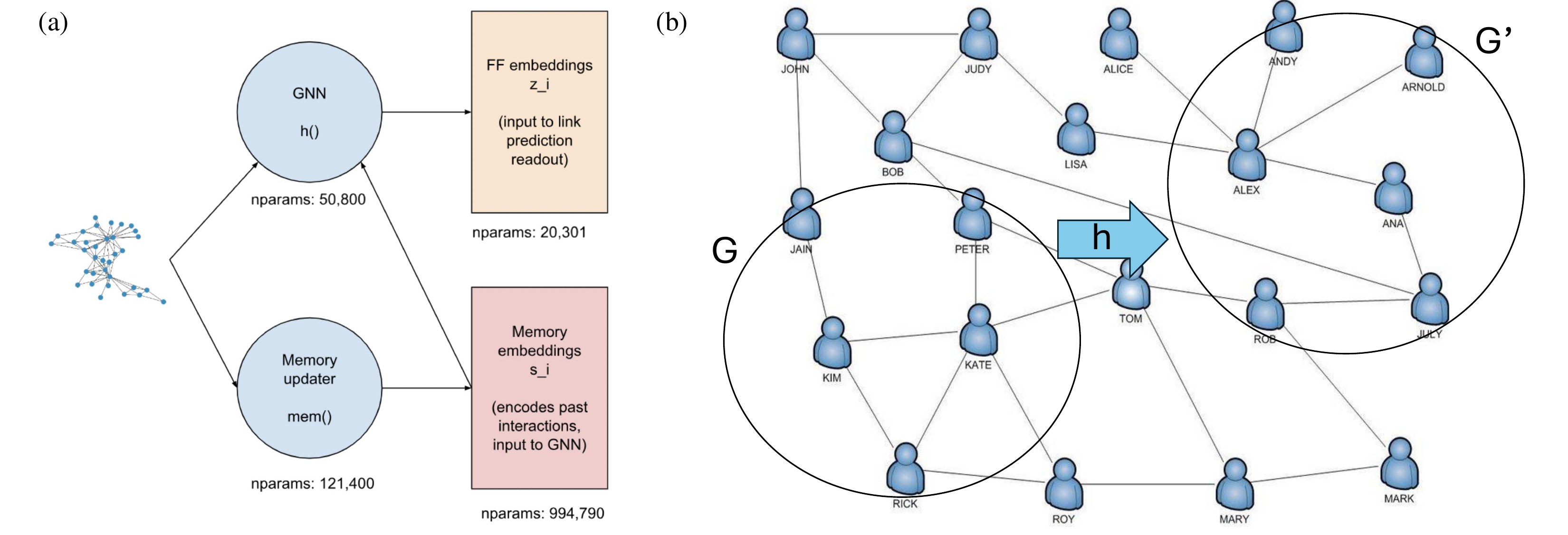}
	\caption{\textbf{The majority of the trained parameters in TGN pertain to the training graph and are not transferable.} \textbf{(a)} Representation of different components of the TGN model trained on \texttt{tgbl-comment} dataset. Here, blue circles are trainable, whereas the other blocks are intermediate state embeddings. We observe that the majority of the variable parameters ($\sim1M$) are associated with the dataset, and only a small fraction of parameters ($\sim150k$) pertain to the model architecture. Hence, the trained TGN model is highly specific to the training data and inherently is less transferable. \textbf{(b)} Transfer learning task for temporal link prediction. We want to train a TLP model on $G$, a community of the larger network, and transfer it to a disjoint community $G'$.}
	\label{fig:parameter_division}
\end{figure*}

In Figure \ref{fig:parameter_division}, we visualize the five TGN components described above, sized according to a training run on the \texttt{tgbl-comment} dataset (consisting of a large batch of Reddit interactions) from the TGB benchmark \cite{huang2023temporal}. The blue components (GNN and memory updater) pertain to the overall model architecture: the parameters of the functions responsible for aggregating and transforming node embeddings. On the other hand, the remaining blocks, representing memory embeddings and GNN embeddings, store parameters that are specific to objects (nodes) seen at training time. This breakdown illustrates that the majority of the learnable parameters in a TGN model are pertained to the representation learning space specific to the training graph, in this case \texttt{tgbl-comment}. This biases the TGN model toward transductive and semi-inductive link predictions --- that is, node pairs where at least one node has been observed at training time -- because the model can pull from the memory store the embedding for the seen node. Hence, for fully-inductive examples where both nodes are new, or in transfer learning scenarios when a new graph with all unseen nodes is introduced, the memory store must be augmented with zero-vectors for the new nodes, which are then updated after the test-time prediction has been made.

We argue that this limitation is inherent to TGN models. In fact, we show that the number of parameters optimized for the training set scales with the number of nodes in the training set, denoted as $N$. The memory store has size \( N \times d_M \) whereas the memory store update MLP has input dimension \( d_M + d_E \) (memory + edge features) and hidden layers \( h_1, h_2, \dots, h_L \), contributing 
\[
(d_M + d_E) \times h_1 + \sum_{i=1}^{L-1} h_i \times h_{i+1} + h_L \times d_M.
\]
The message function MLP, which processes concatenated node embeddings and edge features, adds
\[
(2d_N + d_E) \times h_1 + \sum_{i=1}^{L-1} h_i \times h_{i+1} + h_L \times d_M.
\]
The time encoder has a fixed number of parameters in $N$. Finally, the decoder MLP, which takes concatenated embeddings of two nodes and passes them through hidden layers to output a link score, contributes 
\[
(2d_N) \times h_1 + \sum_{i=1}^{L-1} h_i \times h_{i+1} + h_L \times 1.
\]
Summing all these terms, the total number of parameters in TGN scales like
\[
O\left(N(d_N + d_M) + \max_i{h_i}(\max_i{h_i} + \max\{d_N, d_E, d_M\})\right),
\]
which, when all MLP hidden dimensions are fixed, reduces to $O(N)$.

\subsection{Transfer Learning in Temporal Link Prediction}

The analysis in the last section shows that as the number of nodes $N$ grows large, the amount of information in the TGN model becomes increasingly specific to the training graph nodes, and therefore increasingly unable to be generalized or transferred to new nodes and new graphs. A TGN model trained on the \texttt{tgbl-comment} graph, as illustrated in Fig. \ref{fig:parameter_division}, has $>90\%$ of its parameters devoted to node memory. For the standard temporal link prediction task, where the model is tested on future snapshots of the train graph, this limitation is not too restrictive, because many seen nodes in the training graph participate in new (test-time) interactions. However, in this work, we study how we can transfer the model's knowledge of \emph{temporal interaction dynamics}, i.e., those parameters that are \emph{not} specific to training-time memory, to completely new graphs.

\section{Solution}

In this work, we propose two solutions for enabling TGN models for transfer learning tasks. Specifically, we want to transfer a temporal link prediction model ($h$) trained on graph $G$ to a new graph $G'$ (see Figure \ref{fig:parameter_division}b).

\textbf{Solution 1 - Na\"ive Fine-tuning}: The simplest solution to the TGN transfer learning problem is to fine-tune $h$ trained on $G$ on the first $T_{finetune}$ timesteps of the test graph $G’$. This allows the model to non-trivially initialize some amount of memory store for the unseen nodes. We set $T_{finetune}$ to be the timestamp of the $20\%$ mark through the test set. We then compute test metrics for the remaining $80\%$ of the test data.

\textbf{Solution 2 - Structural Map Transfer Learning}: This is a model-based solution to the transfer learning problem, allowing for a zero-shot alternative to fine-tuning. Specifically, during the training period, we install a new module into TGN that learns a map between \emph{node structural features} and the memory embeddings. Then, at test time, we apply the map to initialize the memory embeddings of unseen nodes. We motivate and describe this approach in the next two sections.

\begin{figure*}
	\centering
	\includegraphics[width=1\linewidth]{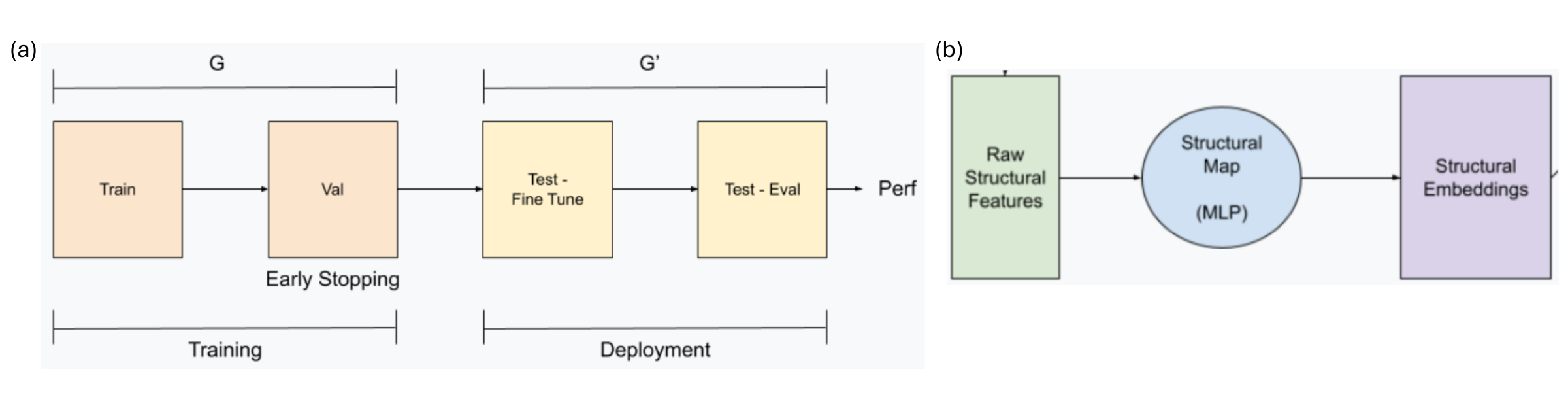}
	\caption{\textbf{Transfer learning framework and structural mapping overview.} \textbf{(a)} We train and validate the TLP model on the training graph $G$. The validation set is used for early-stopping of the model training. During deployment, we learn the memory embeddings of a fraction of the unseen nodes in $G'$ by fine-tuning the model on a fraction of $G'$. This model is thereafter used to derive the test performance on the remaining of $G'$. \textbf{(b)} We learn a structural map during training on $G$. This map learns a function to map the graph topological features to memory embeddings, and when a new node is encountered in the test, we use this learned mapping to initialize the memory embeddings from deterministic topological features.}
	\label{fig:solutions}
\end{figure*}

\subsection{Background on Graph Structural Features}

Ghasemian et al. \cite{ghasemian:pnas2020} showed that graph topological features of nodes, such as degree and centrality, contain information useful for static link prediction. The authors show that stacking simple topological features can achieve sub-optimal link prediction performance across 50 structurally diverse networks from six scientiﬁc domains. In a recent work, He et al. \cite{He2024} have extended the observations to temporal link prediction and have shown the effectiveness of topological features in temporal link prediction. 

Structural features could replicate/improve temporal link prediction performance.
These features provide a common latent space for transfer learning.
We consider multiple topological features for the nodes (degree, betweenness centrality, closeness centrality, and local clustering coefficient), node positional embeddings \cite{yan2024pacer}, orbit counts (substructure counts) \cite{bouritsas2021improving} including cycle graphs, path graphs, complete graphs, binomial trees, star graphs, and non-isomorphic trees.
In our current approach, we utilize the following structural features: 4 topological features (degree, betweenness centrality, closeness centrality, and local clustering coefficient), and node positional embeddings. While orbit counts and other higher-order structural features could provide valuable insights, they are not included in our analysis due to their computational scalability limitations. Given the large scale of our datasets, computing orbit counts is impractical, so we focus on features that are more efficient to compute while still capturing key structural properties. Future work could explore approximate or scalable methods for incorporating such features, possibly by the incorporation of Graph Reservoir Computing \cite{10003110} and efficiently representing the dynamic graph information in a format suitable for the reservoir.

We study the shared information between memory embeddings and these structural features.
The shared information is measured by comparing the node pair-wise distances from memory embeddings and topological features.
We observe a high correlation between the pairwise distances
(e.g., $r_{Pearson}$=0.25 and $r_{Spearman}$=0.42 for \texttt{tgbl-wiki}, $r_{Pearson}$=0.17 and $r_{Spearman}$=0.20 for \texttt{tgbl-flight}).
These observations motivate an approach that learns a map from structural features to memory embeddings and uses the learned map to initialize the memory embeddings of newly observed nodes during transfer learning.

\subsection{Structural Mapping via Temporal Subgraph Aggregation}

\begin{figure*}
	\centering
	\includegraphics[width=1\linewidth]{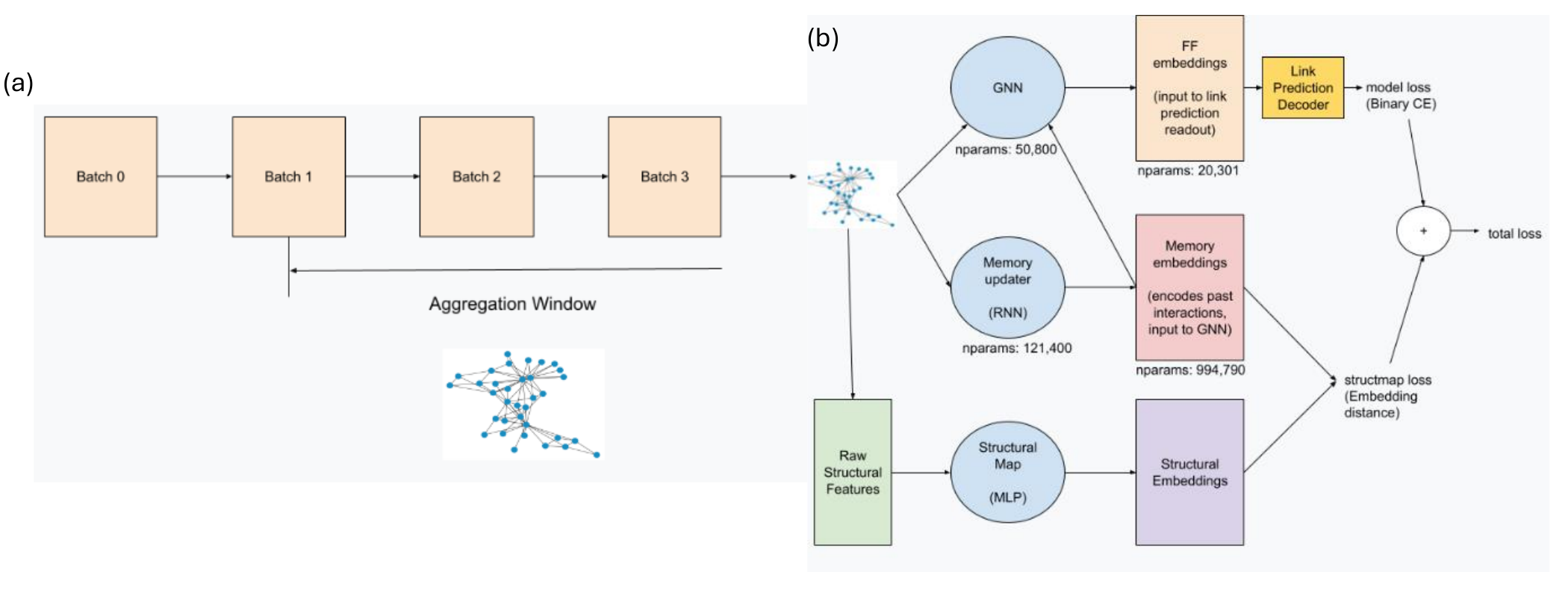}
	\caption{\textbf{Temporal aggregation and structural map architecture} \textbf{(a)} For each newly observed node in test, we aggregate the past edges to construct an aggregated graph, which is used for computing the topological features of the newly incoming node. We use $1\%$ of the time-span of the benchmark datasets as an aggregation window in our experiments. \textbf{(b)} Overview of the structural map-augmented TGN architecture. We use a 3-layer perceptron (MLP) as the structural map module. We combine the loss from the link prediction decoder and the structural map module and train the model in an end-to-end fashion.}
	\label{fig:struct_map}
\end{figure*}

We now describe our solution to the TLP transfer learning problem via graph structure mapping. For each edge batch, we aggregate a window of edges that appeared immediately before the batch, using a window size of $1\%$ of the total time span of the training set.  Structural features, such as node degree and clustering coefficient, are then computed for each node in the aggregated graph. These structural features are fed into the structural map module during training, and an MLP learns the mapping from these structural features to the memory embeddings learned by TGN.
During validation and test, when a new node is encountered, the structural map is used to predict and initialize the memory embeddings for the node (see Figure \ref{fig:struct_map}b). This effectively operates as a ``cold start'' for unseen nodes, allowing for immediate transfer learning without an initial fine-tuning period.

\section{Experiments}

\subsection{Study Design}
We now describe the design of experiments used to test our two approaches.

\subsubsection{Datasets}
We use the link prediction datasets from Temporal Graph Benchmark for the experiments \cite{huang2023temporal}. To set up a transfer learning scenario for each dataset, we divide the graph into two disjoint graphs (with disjoint node sets), using one for training, and the other for validation and testing.
For \texttt{tgbl-wiki} and \texttt{tgbl-review}, we use Louvain community detection \cite{Blondel2008} to divide the nodes, making sure that the community clusters for train, validation, and test are roughly balanced in terms of the number of nodes and time intervals with respect to the original graph dataset and among each other.
For the \texttt{tgbl-flight} dataset, we create train $+$ validation and test communities based on the airports on different continents.

\subsubsection{Evaluation}

We evaluate the test loss across three transfer scenarios:  
\begin{itemize}

    \item \textbf{Transfer without warm-start}: The TGN model, trained solely on the train graph, is applied directly to the test graph without any adaptation.  

    \item \textbf{Transfer with warm-start}: Before deployment, we fine-tune the trained TGN model on an initial temporal segment of the test graph to adapt to new patterns.  

    \item \textbf{Transfer with structural mapping}: Instead of fine-tuning, we initialize the memory embeddings of newly arrived nodes in the test graph using structural mapping while deploying the trained TGN model.  

\end{itemize}

\textbf{Hypotheses:}
\begin{itemize}

    \item \textbf{H1}: We expect the test loss in the \textbf{transfer with warm-start} scenario to be lower than in the \textbf{transfer without warm-start} scenario, demonstrating the benefit of fine-tuning before deployment in a transfer learning setting.  

    \item \textbf{H2}: We anticipate that the test loss in the \textbf{transfer with structural mapping} scenario will be lower than in the \textbf{transfer without warm-start} scenario and comparable to or lower than in the \textbf{transfer with warm-start} scenario. This would validate structural mapping as an effective and computationally efficient alternative to warm-starting.  

\end{itemize}

While we primarily use the loss function to assess the effectiveness of our proposed transfer learning approaches, we plan to expand our evaluation to include additional temporal link prediction metrics, such as Hits@TopK and Mean Reciprocal Rank (MRR) \cite{huang2023temporal}.  

\subsection{Results}

\subsubsection{Fine-tuning}

\begin{figure*}
	\centering
	\includegraphics[width=1\linewidth]{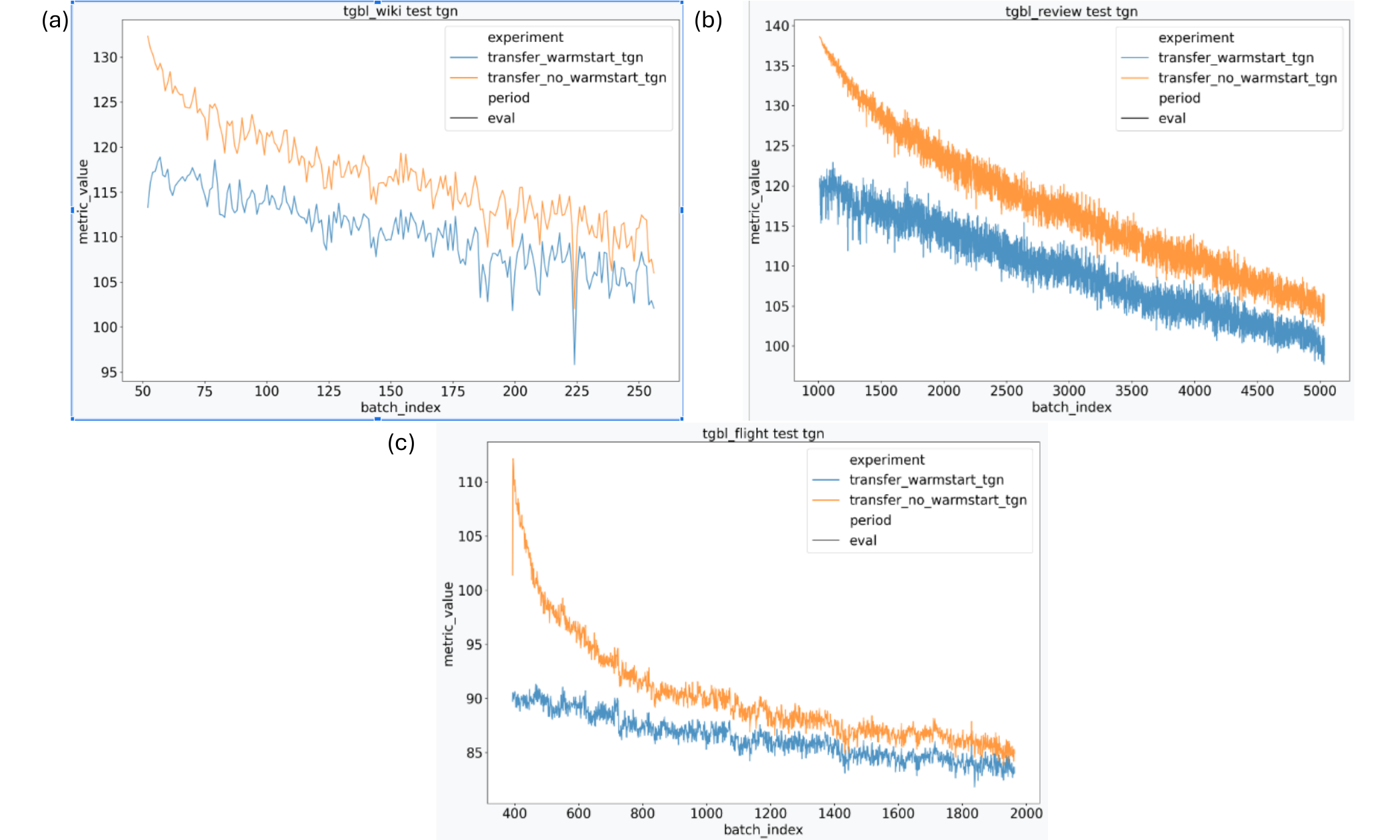}
	\caption{\textbf{Fine-tuning on a fraction of the test graph improves loss during deployment.} We observe lower loss when fine-tuning is implemented in TGN during transfer learning. The improvement achieved by fine-tuning is consistent across multiple benchmark datasets, including \texttt{tgbl-wiki}, \texttt{tgbl-review}, and \texttt{tgbl-flight}. Here, metric value refers to the total loss of the TGN model.}
	\label{fig:finetuning}
\end{figure*}

We fine-tune the trained TGN model on the initial $20\%$ batches of edges of the test data.
We observe lower loss when fine-tuning is applied during transfer learning (see Figure \ref{fig:finetuning}).
However, we would like to avoid fine-tuning since it prevents instant deployment on a new graph, and as a solution, we replace fine-tuning with a zero-shot structural mapping approach.

\subsubsection{Structural Map}

\begin{figure*}
	\centering
	\includegraphics[width=1\linewidth]{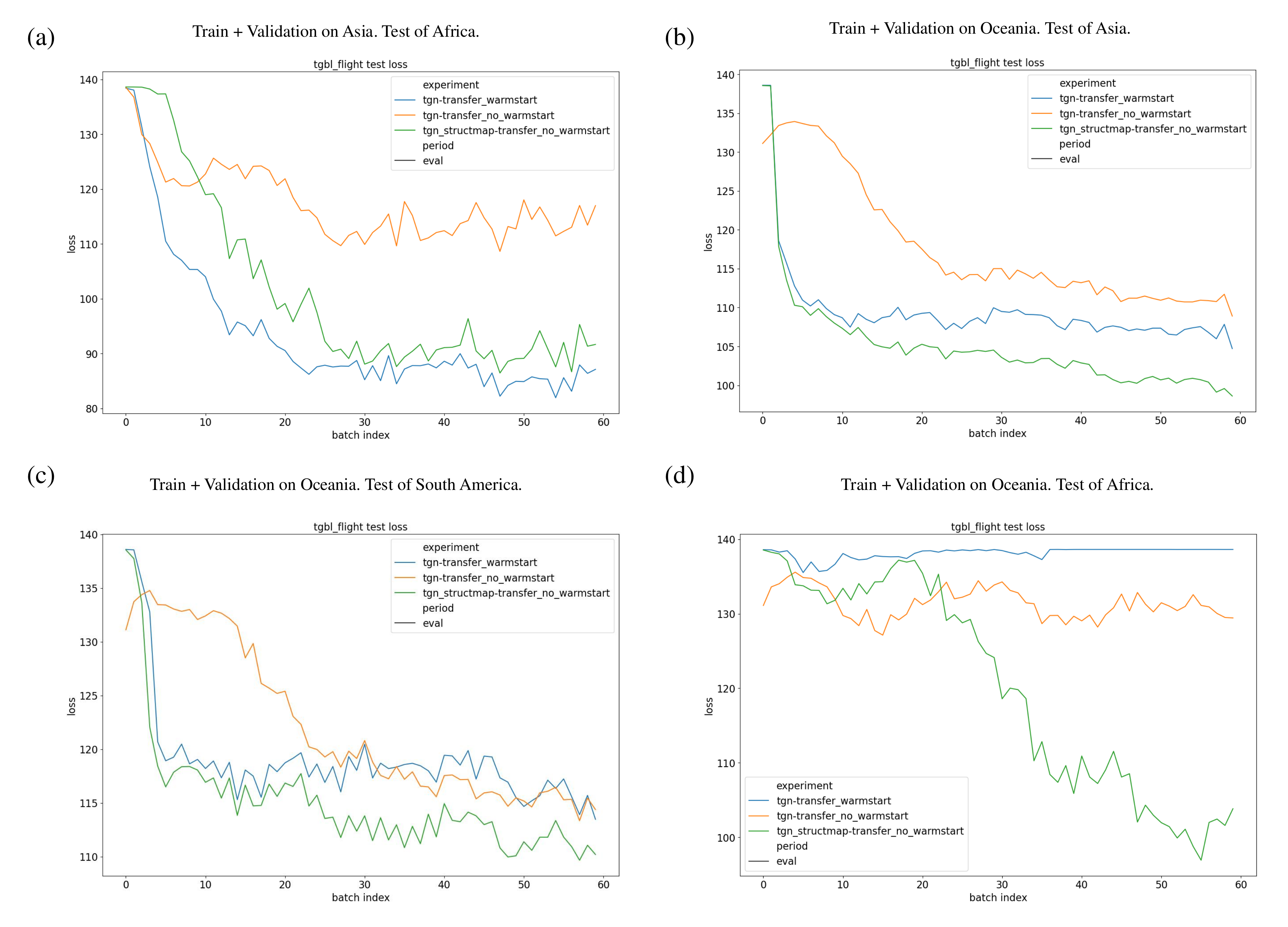}
	\caption{\textbf{TGN with structural mapping improves transfer loss during deployment on an unseen temporal graph.} The \texttt{tgbl-flight} dataset is tested in different transfer learning scenarios. TGN, TGN with fine-tuning, and TGN Structural Map are used in this study. The train and validation graphs pertain the flights and the airports contained in a certain continent, whereas the test graphs are derived from a continent different from the ones used in train and validation. This ensures that the airports trained and validated on are disjoint from the airport tested on. We show that TGN Structural Map can achieve similar or lower model loss compared to the fine-tuned version on the test dataset. The observations are consistent across 4 different transfer learning scenarios.}
	\label{fig:tlp_results}
\end{figure*}

The experiments on the \texttt{tgbl-flight} dataset reveal key insights into the performance of TGN, TGN with fine-tuning, and TGN Structural Map under different transfer learning scenarios. Training and validation graphs are derived from flights and airports within a specific continent, while test graphs are based on a different continent, ensuring that the training and validation airports remain disjoint from those in the test set. Notably, TGN Structural Map demonstrates comparable or lower model loss than the fine-tuned TGN on the test dataset, consistently across all four transfer learning scenarios. This highlights the robustness of the TGN Structural Map in handling domain shifts between training and testing environments (see Figure \ref{fig:tlp_results}).

\section{Limitations}

\subsection{Divergent Model Loss}

\begin{figure}
	\centering
	\includegraphics[width=1\linewidth]{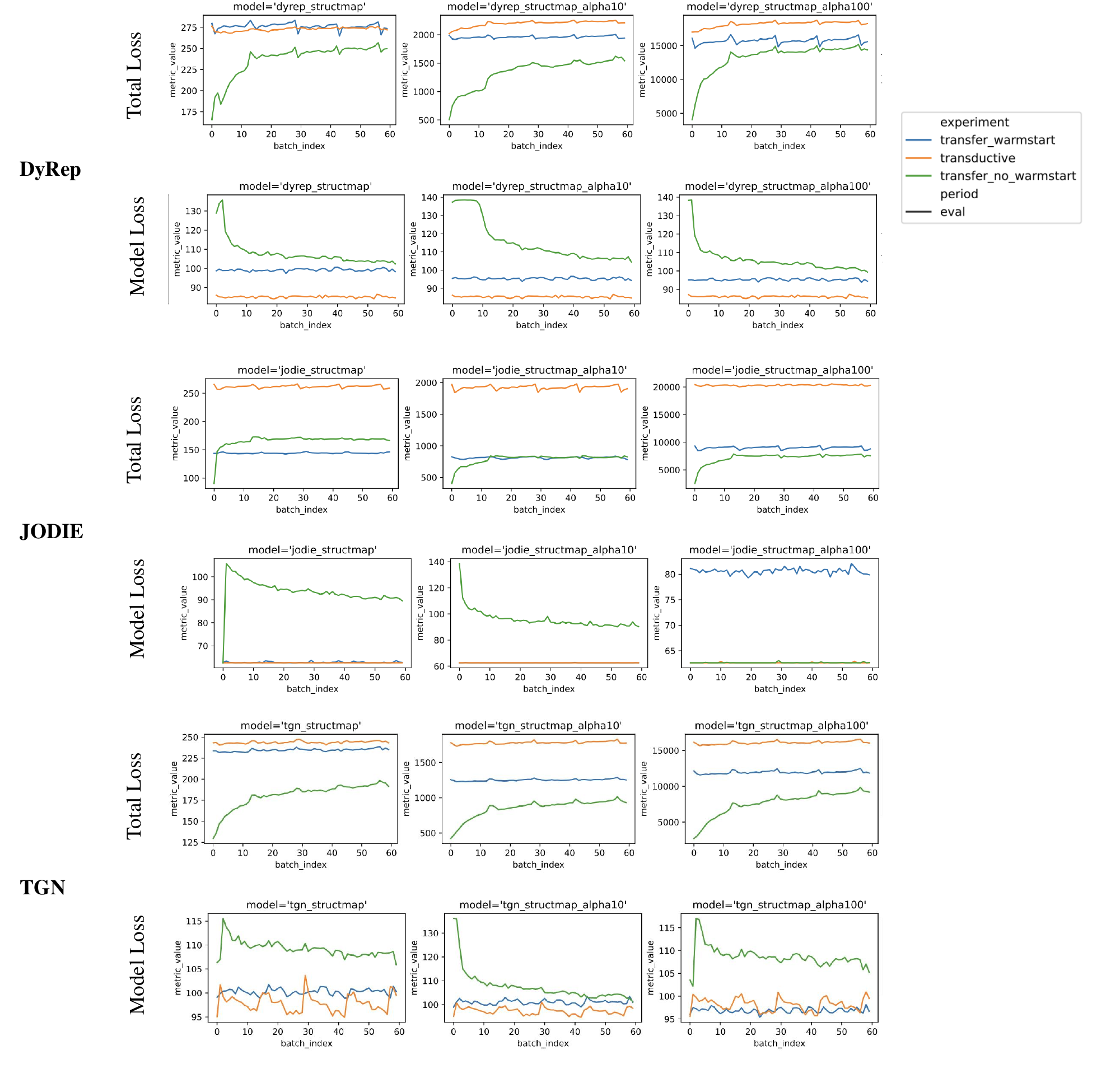}
	\caption{\textbf{Divergent Loss Trends in Transfer Learning: TLP Models vs. Structural Mapping Module.} In a transfer learning setting on \texttt{tgbl-flight} from SA (South America) to AS (Asia), the total loss (i.e., the sum of the TLP model loss and the structural mapping loss) (green lines - first row) increase over train and test edge batches, while the TLP model-only model loss in presence of structural mapping (green line - second row) steadily decreases. This divergent behavior in the first scenario indicates that the structural mapping module (MLP) is not converging effectively during training and deployment. Similar observations persist across other transfer learning scenarios in \texttt{tgbl-flight}, \texttt{tgbl-wiki}, \texttt{tgbl-review}, and \texttt{tgbl-coin}. In the experiments, $\alpha$ represents the weight of the StructMap loss in the total loss, i.e., total loss = $\alpha$ * StructMap loss + TLP model loss.}
	\label{fig:incr_loss}
\end{figure}

During deployment, we frequently observe an increase in the total loss, which comprises both the temporal link prediction (TLP) model loss (e.g., DyRep \cite{trivedi2018dyrep}, JODIE \cite{kumar2019predicting}, or TGN \cite{tgn_icml_grl2020}) and the structural mapping loss. For example, in a transfer learning scenario on \texttt{tgbl-flight}, when transferring from SA (South America) to AS (Asia), the structural mapping loss (green lines in the first row) consistently increases across edge batches during both training and testing, while the TLP model-only loss (second row) steadily decreases (see Figure~\ref{fig:incr_loss}). This divergence suggests that the structural mapping module (MLP) is not converging effectively. Addressing this issue will require targeted hyperparameter optimization of the StructMap module and exploration of alternative architectural designs. Further investigation into the selection of more informative structural features, as well as a deeper analysis of the temporal dynamics of memory embeddings and structural representations, would also be beneficial.

\subsection{Sensitivity to Seed Selection}

\begin{figure}
	\centering
	\includegraphics[width=1\linewidth]{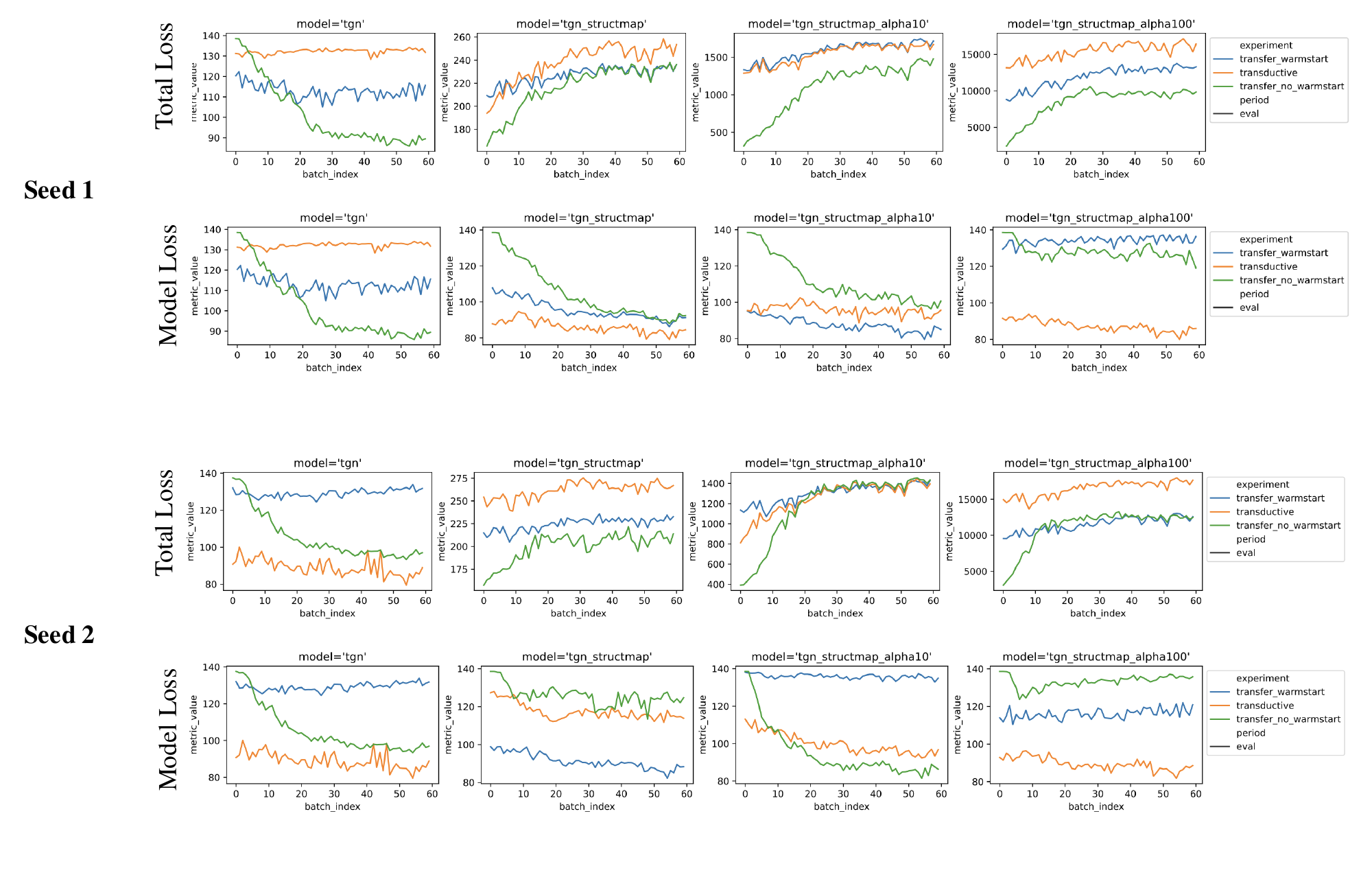}
	\caption{\textbf{Sensitivity of Loss to Random Seed in Transfer Learning.} In a transfer learning setting on \texttt{tgbl-flight} from Asia (AS) to AF (Africa), we observe that both the total loss (i.e., the sum of the TGN model loss and the structural mapping loss) (green lines) and the TGN-only loss (blue lines) exhibit substantial variation when the random seeds for \texttt{torch} and \texttt{numpy} are changed. This highlights potential concerns regarding the robustness and stability of the TGN model and underscores the need for further investigation into TLP model consistency.}
	\label{fig:seed_loss}
\end{figure}

Here, we investigate the impact of random seed selection and initialization on memory embeddings, as well as on the overall behavior of the TGN model and StructMap module. Our observations indicate that both the standalone TGN and its StructMap-augmented variants are highly sensitive to the initial seeds set in PyTorch and NumPy (see Figure \ref{fig:seed_loss}). This sensitivity raises concerns about the robustness and reproducibility of these models, warranting further examination. Studying the temporal evolution of the learned memory embeddings may provide valuable insights into the patterns captured by the TGN model and help inform the selection of structural features for more effective transfer across heterogeneous network domains.

\section{Conclusion and Future Work}

In this work, we examined the impact of temporal fine-tuning on a new test graph and then introduced structural mapping as an alternative. Structural mapping facilitates zero-shot deployment of temporal link prediction models in transfer learning scenarios. 
Our results showed improved deployment test loss for the TGN model across various inter-continental transfer learning scenarios on the benchmark \texttt{tgbl-flight} network. However, further investigation is needed to identify the limitations of structural mapping and explore ways to address them, enabling its applicability to other inter-continental scenario combinations within \texttt{tgbl-flight} and other TGB benchmark datasets, such as \texttt{tgbl-wiki}, \texttt{tgbl-review}, and \texttt{tgbl-comment}.

Associative memory \cite{pmlr-v48-danihelka16} can enhance machine learning models for link prediction in temporal graphs by efficiently capturing and utilizing historical interactions. While models like Temporal Graph Networks (TGNs) integrate memory modules to store and update node embeddings over time, enabling predictions that account for both temporal evolution and structural dynamics, JODIE \cite{kumar2019predicting} employs embedding trajectories and associative memory-inspired mechanisms to adapt to recurrent interactions in dynamic graphs. We plan to develop structural mapping counterparts for associative memory-laden models like JODIE to validate the universality of structural mapping in temporal link prediction models.

Furthermore, persistent homology, a key concept in topological data analysis (TDA), has shown promise in applications to temporal graph learning and link prediction. This method captures the topological features of data, such as connected components, loops, and voids, that persist across different scales. In temporal graphs, persistent homology can help analyze the dynamic evolution of connections and identify robust structural patterns critical for predicting links over time. For instance, persistent homology has been used to extract topological features from subgraphs that evolve over time, enhancing the interpretability and performance of link prediction models. These features are particularly useful in understanding the multi-hop paths and structural connectivity between nodes, as demonstrated by Yan et al., where extended persistent homology was employed to encode rich multi-scale topological information for link prediction tasks \cite{Yan2021LinkPW}.
Similarly, the PHLP method introduced the concept of persistence images to analyze dynamic graph substructures, enabling the integration of diverse topological signals for better predictive power \cite{You2024PHLPSP}.
Integrating persistent homology in learning a transferable map for the memory embeddings could make the transfer learning task less parameter-intensive and improve computation time and resource efficiency.

Finally, Dall’Amico et al. \cite{DallAmico2024} have recently developed distance measures on temporal graphs that cluster temporal graphs. These graph clusters could provide us with insights into the transferability between temporal graphs and guide us toward developing a temporal graph foundation model.

\section*{Code availability}

The codes that support the findings of this study are openly available on the Anonymous GitHub at 
\url{https://github.com/google-research/google-research/tree/master/fm4tlp}.

\section*{Author Contributions}

A.C. contributed to conceiving the project, conducting the literature study, designing the experimental system, developing the structural map model, and writing the manuscript. B.I. contributed to implementing multiple temporal link prediction models, improving the structural mapping module, and writing the manuscript.
B.R. contributed to the literature study, running multiple experiments, and writing the manuscript. J.P. contributed to mentoring the project, conceiving it, conducting the literature study, designing the experimental system, developing the structural map model, and writing the manuscript.

\section*{Acknowledgments}

We want to acknowledge Google DeepMind for allowing us to conceive the project and for providing the necessary resources for its execution. We also thank Northeastern University for providing its Research Computing infrastructure.

\section*{Competing interests}

A.C. is the founder of BioClarity AI LLC., a company utilizing graph machine learning for applications in gene therapy.

\newpage
\bibliographystyle{unsrtnat}
\bibliography{reference}

\end{document}